%% file: paper.tex
\documentclass[hidelinks]{article}

\input{sections/packages}

\title{Gym-preCICE: Reinforcement Learning Environments for Active Flow Control}

\author{Mosayeb Shams\thanks{Corresponding author: m.shams@hw.ac.uk}}
\author{Ahmed H.~Elsheikh}
\affil{Heriot-Watt University, Edinburgh, United Kingdom}

\let\oldmaketitle\maketitle
\renewcommand{\maketitle}{\oldmaketitle\setcounter{footnote}{0}}

\begin{document}

\maketitle

\begin{abstract}
\input{sections/abstract}
\end{abstract}


\section{Motivation and significance}
\label{sec:motivation}
\input{sections/introduction}

\section{Software description}
\label{sec:software}
\input{sections/software}

\section{Illustrative Examples}
\label{sec:examples}
\input{sections/examples}
\section{Impact}
\label{sec:impact}
\input{sections/impact}

\section{Conclusions}
\label{sec:conclusions}
\input{sections/conclusions}

\section*{Acknowledgements}
This work was supported by the Engineering and Physical Sciences Research Council grant number EP/V048899/1.

\bibliographystyle{unsrtnat}

\end{document}

%% file: sections/packages.tex
\usepackage{arxiv}

\usepackage[utf8]{inputenc} 
\usepackage[T1]{fontenc}    
\usepackage{hyperref}       
\usepackage{url}            
\usepackage{booktabs}       
\usepackage{amsfonts}       
\usepackage{nicefrac}       
\usepackage{microtype}      
\usepackage{graphicx}
\usepackage{doi}
\usepackage{authblk}
\usepackage{float}
\usepackage{varwidth}
\usepackage{bm}
\usepackage[english]{babel}
\usepackage{mathptmx} 
\usepackage{tikz-dimline}
\usepackage{amsmath}
\usepackage{amssymb}
\usepackage[obeyDraft, obeyFinal]{todonotes}
\usepackage{bm}
\usepackage{tcolorbox}
\usepackage{subfiles}
\usepackage{colortbl}
\usepackage{ctable}
\usepackage{etex}
\usepackage{paralist}
\usepackage{wasysym}
\usepackage{subcaption}
\usepackage{multirow}
\usepackage{pifont}
\usepackage{wrapfig}
\usepackage{placeins}
\usepackage{acronym}
\usepackage{paralist}

\usepackage[numbers]{natbib}

\usepackage[frozencache]{minted}
\setminted{linenos, highlightcolor=blue!10, breaklines, fontsize=\small}

\usepackage{tikz}
\usetikzlibrary{calc,tikzmark,decorations.pathreplacing,calligraphy}

%% file: sections/abstract.tex
Active flow control (AFC) involves manipulating fluid flow over time to achieve a desired performance or efficiency. AFC, as a sequential optimisation task, can benefit from utilising Reinforcement Learning (RL) for dynamic optimisation. In this work, we introduce Gym-preCICE, a Python adapter fully compliant with Gymnasium (formerly known as OpenAI Gym) API to facilitate designing and developing RL environments for single- and multi-physics AFC applications. In an actor-environment setting, Gym-preCICE takes advantage of preCICE, an open-source coupling library for partitioned multi-physics simulations, to handle information exchange between a controller (actor) and an AFC simulation environment. The developed framework results in a seamless non-invasive integration of realistic physics-based simulation toolboxes with RL algorithms. Gym-preCICE provides a framework for designing RL environments to model AFC tasks, as well as a playground for applying RL algorithms in various AFC-related engineering applications.

%% file: sections/introduction.tex
Active flow control (AFC) has been the subject of extensive research in a wide range of fluid engineering applications due to its potential in enhancing performance and efficiency \citep{cattafesta2011actuators,kral2000active,collis2004issues}. Active flow control is the process of targeted manipulation of flow dynamics through exerting a small amount of energy input to accomplish a prescribed objective such as drag attenuation, mixing augmentation, vortex-vibration-induced suppression, heat transfer enhancement, etc. \cite{collis2004issues}. In a closed-loop active flow control system, online information from the flow provided by sensors is used as feedback to adjust the controller to a wide operation envelope as well as incoming perturbations. However, due to the high-dimensionality and non-linearity of fluid dynamics, it is often extremely challenging to design effective and robust control strategies for flow problems using classical active flow control methods \cite{ren2020active}. Alternatively, deep reinforcement learning (DRL) algorithms are well-suited to learn robust control strategies for partially observed systems with high-dimensional nonlinear stochastic dynamics \cite{sutton2018reinforcement}.

Over the last decade, DRL has been demonstrated as a viable approach to solve various control tasks in robotics and video-gaming \cite{Silver2017, schrittwieser2020mastering, Hafner2023}. Within a DRL cycle, a deep neural network called {\it agent} is trained to become an intelligent decision maker via interacting with an {\it environment}, which is often a simulation engine \cite{sutton2018reinforcement}. Recently, Gymnasium, formerly known as OpenAI Gym, has become a de facto standard API to communicate between DRL algorithms and simulation environments \cite{brockman2016openai}. While DRL research in the fields of robotics and video-gaming have greatly benefited from this standardised API, there is a lack of a similar software tools in the field of computational mechanics, specifically computational fluid dynamics (CFD), to facilitate developing and comparing DRL algorithms for AFC problems.

In recent years, a number of researchers have investigated solving AFC problems using DRL \cite{Rabault2019a,rabault2020deep,paris2023reinforcement,Wang2022}. Despite important advances, the published results are often neither directly comparable nor easily extendable due to inflexible designs, inaccessible source codes, or the utilisation of different DRL libraries. This work takes the first step towards a non-invasive open-source unification of widely used and reliable DRL libraries (e.g., Stable-Baselines3 \cite{stable-baselines3}, CleanRL \cite{huang2022cleanrl}, etc.) and open-source simulation toolboxes to enable the rapid development of DRL-based AFC solutions with possible extensions to various engineering control problems.

In this work, we introduce Gym-preCICE, a framework that facilitates the design and the implementation of reinforcement learning (RL) environments for AFC with the special focus on CFD-in-the-loop training procedures. Gym-preCICE couples Gym-style DRL-based algorithms to external mesh-based partial differential equation (PDE) solvers via preCICE \cite{chourdakis2021precice}, an open-source multi-physics coupling library.

Our design choice of using preCICE, which is originally developed to couple numerical solvers for multi-physics simulations, as the coupling interface between DRL algorithms and PDE solvers has been motivated by two main factors: (1) preCICE enables black-box coupling of DRL-algorithms and mesh-based numerical solvers in an agnostic and non-invasive fashion, which is the crucial building block to facilitate rapid adaptation of DRL algorithms in complex physics-based dynamic systems, and (2) preCICE is becoming a popular coupling choice among the researchers in multi-physics due to its large number of ready-to-use coupling adapters that cover a wide range of PDE solvers (e.g., OpenFOAM \cite{weller1998tensorial}, deal.II \cite{arndt2022deal}, FEniCS \cite{alnaes2015fenics}, CalculiX \cite{dhondt2017calculix}, XDEM \cite{peters2019xdem}, etc.). This can allow for easy extension of our DRL training framework to a vast range of multi-physics systems.

For the purposes of demonstration, we use Gym-preCICE to train a state-of-the-art RL algorithm called proximal policy optimisation (PPO) \cite{schulman2017proximal}, for drag attenuation in laminar flow over a cylinder \cite{schafer1996benchmark, Rabault2019a}. We demonstrate the flexibility and modularity of Gym-preCICE by employing two different actuation systems for drag attenuation, namely synthetic jet and rotating-cylinder actuators. Moreover, using a fluid-structure interaction (FSI) control test case, we demonstrate that Gym-preCICE's coupling capabilities are adaptable for both single and multi-solver physics simulation engines. We expect that Gym-preCICE, in particular, will help and inspire the fluid dynamics research community to effectively address research questions concerning intelligent AFC using DRL without being buried in implementation details.

In Section~\ref{sec:software}, we provide a detailed description of the software. In Section~\ref{sec:examples}, we present three examples of how the adapter is used in closed- and open-loop AFC applications. In Section~\ref{sec:impact}, we discuss the impact of the new software.

%% file: sections/software.tex
\begin{figure}[t]
    \centering
    \includegraphics[scale=1.0]{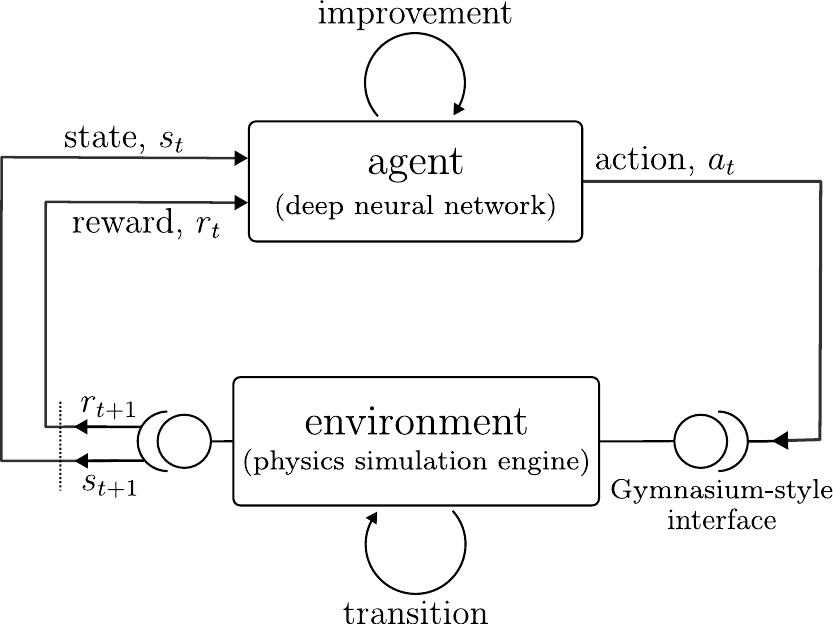}
    \caption{Deep Reinforcement Learning (DRL) cycle: an agent (a deep neural network) is trained to become an intelligent decision maker via interacting with an environment (a physics-based simulation engine). At time step ${t}$, the agent partially observes the state of the environment, ${s_t}$, evaluates its behaviour based on a received reward, $r_t$, and outputs an action, $a_t$, in an attempt to control the environment in a favourable way.}
    \label{fig:1}
\end{figure}

We first define the DRL terminology that we use in this work. We consider a control setting in which a deep neural network called \emph{agent} interacts with an \emph{environment}, which represents a behind-the-scenes physics simulation engine. As shown in Figure~\ref{fig:1}, at each interaction cycle, the agent outputs an action to the environment, and receives observation and reward from the environment \cite{sutton2018reinforcement}. The tuple of action, observation, and reward is called an \emph{experience}, which is utilised by the agent to improve its decision-making capabilities during training. We refer to a sequence of these experience tuples from the start to the end of a simulation as an \emph{episode}. In our work, we only deal with \emph{episodic} tasks with a predetermined terminal condition, and the goal is to achieve the highest level of performance within the fewest number of episodes. The goodness of the action taken by the agent is determined by a scalar signal called \emph{reward}. We use the word \emph{observation} to refer to a set of flow field variables such as pressure, velocity, force, temperature, displacement, etc. probed at predefined locations across the simulation domain, and communicated with the agent at every interaction step. We refer to the set of all valid actions and observations in all states as the \emph{action space} and \emph{observation space}, respectively. 

To understand how Gym-preCICE adapter works, we need to know how it interfaces with other software packages in a control loop. Therefore, first, in Section~\ref{ssec:SoftwareArchitecture}, we give an overview of the software architecture and introduce the packages involved. Then, in Section~\ref{ssec:SoftwareFunctionalities}, we provide detailed information on the adapter's API and methods, along with a brief summary of its parallelisation capabilities and the testing procedures employed to validate the new software.

The software is developed and maintained on GitHub\footnote{\url{https://github.com/gymprecice/gymprecice}}, and its source code is publicly available under the MIT license.

\subsection{Software architecture}
\label{ssec:SoftwareArchitecture}
\begin{figure}[t]
    \centering
    \includegraphics[scale=1.0]{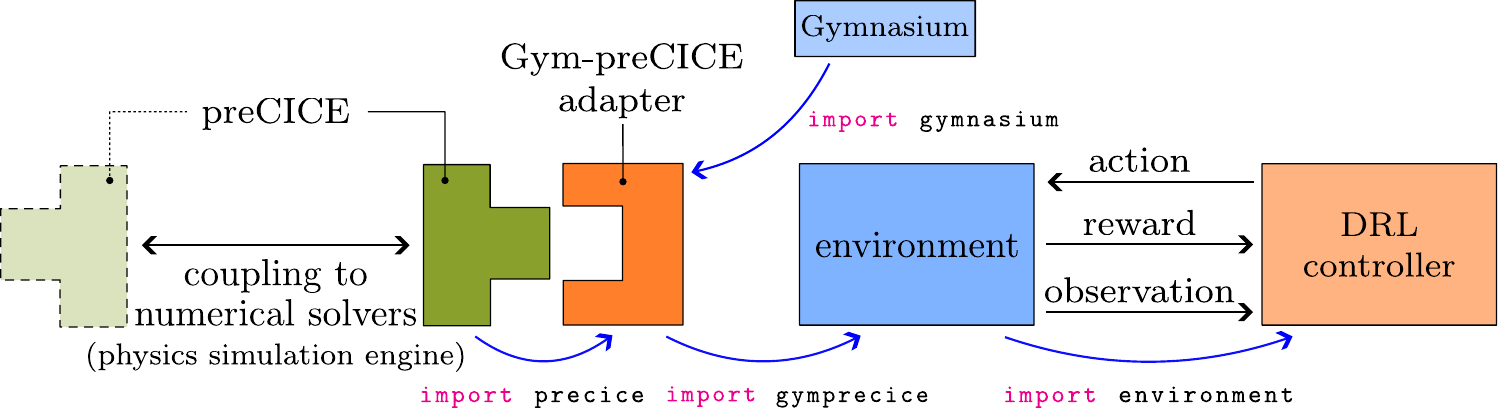}
    \caption{General layout of Gym-preCICE software structure. The original contribution of this work is in developing Gym-preCICE adapter block.}
    \label{fig:2}
\end{figure}

Figure~\ref{fig:2} depicts an overview of constituent elements of the developed software, and how these elements interact with each other in a closed-feedback control loop. The loop consists of three main parts: (1) a DRL controller, (2) an environment, and (3) a single- or multi-solver physics simulation engine. From the preCICE point of view, the controller and the physics simulation engine are participants in a coupled simulation setup rather than a control loop. Therefore, the necessary adjustments are implemented within a middle-layer software called Gym-preCICE adapter to meet both preCICE and DRL expectations. The adapter acts as a glue-code between the controller and the coupling library preCICE, which in turn communicates with the physics simulation engine. In the following, we detail individual blocks of the software layout.

\emph{Gymnasium} is a toolkit for reinforcement learning (RL) research introduced by OpenAI and currently maintained by Farama Foundation\footnote{\url{https://gymnasium.farama.org/}} with the aim of standardising the development of RL algorithms by providing a simple abstract interface capable of representing generic RL problems.

\emph{preCICE} is an open-source coupling library for partitioned multi-physics simulations\footnote{\url{https://precice.org/index.html}}. It is written in C++, but also offers additional bindings for C, Fortran, Python, and MATLAB. preCICE manages communication between mesh-based PDE solvers called participants, which run as independent programs. This offers the modular flexibility needed to setup complex multi-physics simulations. At run-time, preCICE is configured with an \mintinline{bash}{xml} file describing the coupling setup\footnote{\url{https://www.precice.org/configuration-overview.html}}.

\emph{Gym-preCICE adapter} provides an abstract API for coupling reinforcement learning algorithms to PDE solvers within a physics simulation engine via preCICE. The adapter mainly consists of wrapper functions that call preCICE commands, and a set of complementary functions to check whether the physics simulation engine has reached a terminal state, and to reset it between episodes. A detailed explanation of the adapter API is given in Section~\ref{ssec:SoftwareFunctionalities}.

\emph{DRL controller} is an algorithm concerned with a sequential decision-making control task. The controller uses a problem-specific user-provided Python code as its environment and a deep neural network as its agent.

\emph{environment} is a user-provided Python sub-class that inherits from Gym-preCICE adapter and overrides its non-generic abstract methods. It contains problem-specific information such as geometric data, action space, observation space as well as reward, read/write and data conversion functions. The DRL controller uses the environment to communicate with the behind-the-scenes physics simulation engine via preCICE. We present a few problem-specific environments in Section~\ref{sec:examples}.

\subsection{Software functionalities}
\label{ssec:SoftwareFunctionalities}
The Gym–preCICE adapter is designed as a stand-alone and generic adapter to minimise the setup efforts required for users of preCICE and Gym-based RL libraries. Here, we describe main components of the adapter's API in details as shown in Listing~\ref{list:1}.
\begin{listing}[p!]
\begin{center}
\begin{minipage}{0.9\textwidth} 
\begin{minted}[escapeinside=!!]{python}
import gymnasium as gym  # standard Gymnasium interface for reinforcement learning!\label{adapter_depencency_start}!
import precice  # Python bindings of preCICE for multi-physics coupling
[...]  # other dependencies!\label{adapter_depencency_end}!

class Adapter(ABC, gym.Env):  # abstract class that inherits from Gymnasium!\label{adapter_init_start}!
    def __init__(self, options, idx):
        [...]  # setup generic attributes!\label{adapter_init_end}!

    # overridden Gymnasium public methods:
    def reset(self, *, seed, options):!\label{adapter_reset_start}!
        [...]  # reset the base class and close external resources
        self._launch_subprocess(...)  # launch physics simulation engine (PDE solvers)
        self._init_precice(...)  # create and initialise preCICE interface
        [...]  # perform necessary checks on the spawned sub-processes
        return self._get_observation(...), {}  # return initial observation to controller!\label{adapter_reset_end}!
    
    def step(self, action):!\label{adapter_step_start}!
        # map control actions to surface boundary values on actuation interfaces
        write_data = self._get_action(action, self._write_var_list)
        read_data = self._advance(write_data)
        observation = self._get_observation(read_data, self._read_var_list)  
        reward = self._get_reward()
        terminated = self._is_episode_terminated()
        [...]  # if terminated, finalise coupling and prepare to reset environment 
        return observation, reward, terminated, False, {}  # return experience tuple!\label{adapter_step_end}!

    def close(self):!\label{adapter_close_start}!
        [...]  # close environment and release external resources!\label{adapter_close_end}!

    # private methods redirected to preCICE:
    def _init_precice(self):!\label{adapter_precice_start}!
        self._interface = precice.Interface(...)  # create preCICE interface
        [...]  # set spatial mesh coupling data
        self._dt = self._interface.initialize()  # initialise preCICE interface
        [...]  # set read/write mesh coupling data

    def _is_episode_terminated(self):
        terminated = not self._interface.is_coupling_ongoing()
        return terminated
        
    def _advance(self):
        [...]  # write mesh coupling data to preCICE buffer
        self._interface.advance(self._dt)  # advance physics simulation engine's dynamics
        [...]  # read and return mesh coupling data from preCICE buffer!\label{adapter_precice_end}!
    
    # abstract methods to be overridden within problem-specfic AFC environments:
    @abstractmethod!\label{adapter_abstract_start}!
    def _get_action(self, action, self._write_var_list):
        raise NotImplementedError
    
    @abstractmethod
    def _get_observation(self, read_data, self._read_var_list):
        raise NotImplementedError
    
    @abstractmethod
    def _get_reward(self):
        raise NotImplementedError
    
    @abstractmethod
    def _close_external_resources(self):
        pass!\label{adapter_abstract_end}!
\end{minted}
\end{minipage}
\end{center}
\caption{The code excerpt of Gym-preCICE API.}
\label{list:1}
\end{listing}

\paragraph{External libraries (lines \ref{adapter_depencency_start} -- \ref{adapter_depencency_end})}
The adapter relies on Gymnasium and preCICE Python libraries to handle controlling and coupling tasks, respectively.

\paragraph{Adapter definition and initialisation (lines \ref{adapter_init_start} -- \ref{adapter_init_end})} 
The adapter is an abstract Python class that inherits from the Gymnasium base-class, \mintinline{python}{Env}, serving the basis for all problem-specific environments. \mintinline{python}{Env} itself is an abstract class that encapsulates an environment with arbitrary under-the-hood dynamics. Within our adapter, we override three main abstract methods of \mintinline{python}{Env} API: \mintinline{python}{reset}, \mintinline{python}{step}, and \mintinline{python}{close}. This allows us to implement the necessary coupling machinery using preCICE to handle all the communications between the DRL controller and the physics simulation engine, which dictates the dynamics of the environment. The adapter is configured with \mintinline{bash}{options} argument, which is a Python dictionary containing the setup information for the environment, the controller, and the physics simulation engine.

\paragraph{Starting an episode (lines \ref{adapter_reset_start} -- \ref{adapter_reset_end})} 
At the start of each episode, we need to reset the environment, more specifically the physics simulation engine, to an initial state and return a partial observation to the DRL controller. To follow the partitioned coupling framework of preCICE, \mintinline{python}{_launch_subprocess} method spawns separate sub-processes to simultaneously run PDE solvers within the physics simulation engine. We interface Gym-preCICE adapter to preCICE Python bindings by calling private wrapper \mintinline{python}{_init_precice}. The initial observation is returned by calling method \mintinline{python}{_get_observation}.

\paragraph{Steering an episode (lines \ref{adapter_step_start} -- \ref{adapter_step_end})} 
For each interaction step between the agent and the environment within an episode, we run one simulation time step (or multiple simulation time steps within a so-called time-window) of the physics simulation engine's dynamics. preCICE is intended to perform bi-directional surface coupling between two or more mesh-based PDE solvers called participants. In our case, however, one of the participants is a DRL controller, rather than a PDE solver. The controller actively manipulates the simulation domain by controlling boundary values on so-called actuation interfaces. The actuation interfaces are physical boundaries belong to the PDE solvers within the physics simulation engine. Therefore, to be consistent with the workflow of preCICE, we need to transform control actions to appropriate surface boundary values on the actuation interfaces using abstract method \mintinline{python}{_get_action}.  Finally, the time and the coupling loop control are handed over to preCICE by calling wrapper method \mintinline{python}{_advance}, which under-the-hood writes the control surface boundary values to the physics simulation engine, advances its dynamics one step forwards in time, and reads values of non-actuation interfaces from the physics simulation engine. After each step, we get the current partial observation and compute the reward signal, through problem-specific abstract methods, and check whether the end of the episode is reached before finally returning the observation-reward-terminated tuple to the DRL controller.

\paragraph{Closing (lines \ref{adapter_close_start} -- \ref{adapter_close_end})} 
Any necessary clean-up of external resources is performed upon concluding a control task.

\paragraph{Redirecting to preCICE (lines \ref{adapter_precice_start} -- \ref{adapter_precice_end})} 
The preCICE functionalities required by Gym-preCICE API public methods, namely \mintinline{python}{reset}, \mintinline{python}{step}, and \mintinline{python}{close}, are all invoked via private wrapper methods. These methods take care of creating the main access point to a preCICE interface object, exchanging data with the PDE solvers within the physics simulation engine as well as controlling and monitoring the time and the coupling loop during an episode.

\paragraph{Handling problem-specific operations (lines \ref{adapter_abstract_start} -- \ref{adapter_abstract_end})} 
Gym-preCICE adapter provides four abstract methods to be overridden within a problem-specific Python class defined by users. The DRL controller employs the problem-specific class as its \mintinline{bash}{environment} for interaction purposes. The method~\mintinline{python}{_get_action} allows users to translate actions received from the DRL controller into appropriate surface boundary values on the actuation interfaces. The method~\mintinline{python}{_get_observation} allows users to process surface boundary values, data fields, or data files generated by the physics simulation engine to extract partial observation information required by the DRL controller. The method~\mintinline{python}{_get_reward} allows users to formulate a scalar reward signal required by the DRL controller. The method~\mintinline{python}{_close_external_resources} allows users to close and clean pre-and post-processing files and resources utilised within each episode. 

\paragraph{Parallelisation.}
Gym-preCICE adapter allows multi-environment training where the DRL controller collects data in parallel from several environments, each of which being an independent self-sufficient physics simulation engine. Moreover, the adapter can directly support multi-process simulations by running a parallel PDE solver via the usual \mintinline{bash}{mpirun} command, provided that the PDE solver and its assigned preCICE adapter support distributed-memory parallelisation based on the message-passing interface (MPI).

\paragraph{Testing.}
We use \mintinline{python}{pytest} to comprehensively test Gym-preCICE adapter and its utilities to ensure that it performs as expected under a variety of conditions, increasing its reliability and robustness. To facilitate fast, reliable and reproducible testing of the adapter, we treat it as an isolated unit that is independent of preCICE and any coupled physics simulation engine \cite{rodenberg2021fenics}. For this purpose, first, we replace the preCICE dependency with a mocked version. Then, we use \mintinline{python}{pytest-mock}\footnote{ \url{https://pytest-mock.readthedocs.io/en/latest/index.html}}, which provides a set of fixtures and utilities for testing with mock objects, to simulate the behaviour of preCICE and other related external API calls within our individual unit test functions.

%% file: sections/examples.tex
In this section, we demonstrate the usage and advantages of Gym-preCICE adapter by employing it for three different control scenarios, each of which exemplifies different beneficial characteristics of the adapter. In the first example, we couple a DRL controller to an OpenFOAM fluid solver to show the application of the adapter for AFC. In the second example, we repeat the same scenario as in the first example but using a different flow control actuator to demonstrate the flexibility and modularity of the adapter. In the third example, to showcase the capability of the adapter in handling multi-solver physics simulation engines, we couple a predefined sinusoidal controller to a fluid–structure interaction (FSI) environment, where a deal.II solid solver is coupled to an OpenFOAM fluid solver. For these examples, we use Gym–preCICE adapter release v0.1.0\footnote{\url{https://github.com/gymprecice/gymprecice/releases/tag/v0.1.0}}, along with Gymnasium release v0.28.0\footnote{\url{https://github.com/Farama-Foundation/Gymnasium/releases/tag/v0.28.0}}, preCICE release v2.4.0\footnote{\url{https://github.com/precice/precice/releases/tag/v2.4.0}}, the Python bindings release v2.5.0.1\footnote{\url{https://github.com/precice/python-bindings/releases/tag/v2.5.0.1}}, OpenFOAM-preCICE adapter release v1.2.0\footnote{\url{https://github.com/precice/openfoam-adapter/releases/tag/v1.2.0}}, deal.II-preCICE adapter\footnote{\url{https://github.com/precice/dealii-adapter}}, OpenFOAM release v2112\footnote{\url{https://gitlab.com/openfoam/openfoam/-/tags/OpenFOAM-v2112}}, and deal.II release v9.4.2\footnote{\url{https://github.com/dealii/dealii/releases/tag/v9.4.2}}. 

\subsection{Closed-loop active flow control}
\begin{figure}[t!]
    \centering
    \includegraphics[scale=1.0]{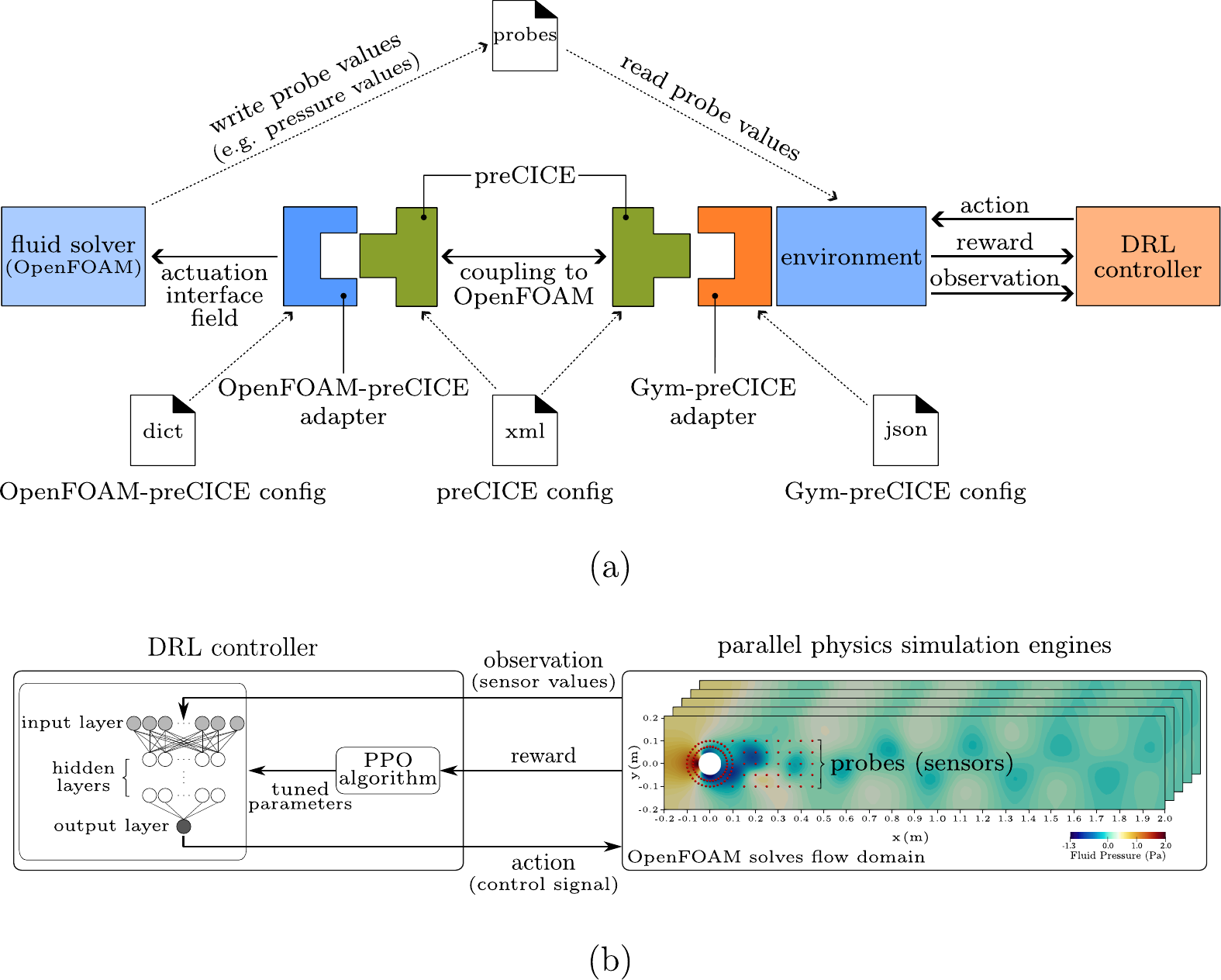}
    \caption{Schematics of (a) software architecture for DRL-based CFD-in-the-loop AFC using OpenFOAM, and (b) closed-loop training framework in which a deep neural network is optimised using the proximal policy optimisation (PPO) algorithm to become an optimal decision-making agent via interacting with multiple OpenFOAM fluid flow simulations (parallel simulation engines).}
    \label{fig:3}
\end{figure}

In the first two examples, we illustrate CFD-in-the-loop DRL-based AFC using Gym-preCICE adapter as its main area of application. Figure~\ref{fig:3}a shows the general software structure where Gym-preCICE adapter is used to couple a DRL controller and OpenFOAM CFD library. Figure~\ref{fig:3}b schematically depicts the closed-loop training framework where OpenFOAM fluid flow simulations (parallel simulation engines) are coupled with the proximal policy optimisation (PPO) algorithm, via Gym-preCICE adapter, to train a deep neural network as the optimal decision-making agent for AFC. The process of the closed feedback control loop is as follows: (1) a problem-specific environment receives a control input (action) from the DRL controller, (2) the environment converts the control input to appropriate surface boundary values (actuation interface fields), and passes them to Gym-preCICE adapter, (3) the adapter writes the boundary values to a buffer with the help of preCICE, (4) OpenFOAM-preCICE adapter reads the boundary values from the buffer and passes them to the OpenFOAM solver \cite{chourdakis2023openfoam}, (5) the solver updates its internal state based on the new boundary values, and writes a new state (observations) into so called probe files, and (6) the environment reads the observation from probe files, computes a reward signal based on the received information, and feeds back the observation alongside the reward to the DRL controller.

\subsubsection{Synthetic jet flow control}
As a simple DRL-based AFC test case, we consider drag reduction in two-dimensional incompressible flow over a rigid cylinder achieved through controlling a pair of small jets on the cylinder\footnote{\url{https://github.com/gymprecice/tutorials/tree/v0.1.0/closed_loop_AFC/jet_cylinder}} \cite{Rabault2019a}.

As shown in Listing~\ref{list:2}, new AFC environments can be implemented as Python sub-classes that inherit from Gym-preCICE adapter. The sub-class API is simple and generic, which can serve as a template to define environments for various AFC problems. To handle operations that are unique to an AFC problem and not generic, the sub-class overrides the abstract methods of the inherited \mintinline{bash}{Adapter} super-class.  Environment parameters such as geometric data, number of probes, external files, action space, observation space, etc., are defined in the environment's constructor, \mintinline{bash}{__init__}. Line \ref{env_get_action} overrides \mintinline{bash}{_get_action} abstract method to convert a flow rate control action received from the DRL controller to parabolic velocity fields on jet actuators. Line \ref{env_get_observation} overrides \mintinline{bash}{_get_observation} abstract method to read pressure probe data from a file and convert them to appropriate input values used as observation by the DRL controller. Line \ref{env_get_reward} overrides \mintinline{bash}{_get_reward} abstract method to formulate a reward signal to guide the DRL controller towards minimising drag on the cylinder while penalising large lift forces. Line \ref{env_close} overrides \mintinline{bash}{_close_external_resources} abstract method to close and reset external resources used by the environment during an episode. Furthermore, line \ref{env_step} overrides \mintinline{bash}{step} public method to enforce the constrain that the PPO agent can interact with the simulation and modify its control only once every 50 numerical time steps, while ensuring that the control action is smoothly and linearly distributed over these steps \cite{Rabault2019a}.
\begin{listing}[!ht]
\begin{center}
\begin{minipage}{0.9\textwidth} 
\begin{minted}[escapeinside=!!]{python}
from gymprecice.core import Adapter  # provide Gym-preCICE adapter as the super-class
[...]  # other dependencies 

class JetCylinderEnv(Adapter):  # new AFC environment that inherits from Adapter
    def __init__(self, options, idx):
        super().__init__(options, idx)  # initialise Adapter
        [...]  # configure environment (geometric data, action-observation spaces, etc.)
        [...]  # extract spatial coordinates of actuation and observation interfaces
        self._set_precice_vectices(interface_coords)  # pass the coordinates to Adapter

    # map control action to surface boundary fields applied on actuation interfaces
    def _get_action(self, action, write_var_list):!\label{env_get_action}!   
        [...]

    # map probes and/or surface boundary fields to an observation input for controller
    def _get_observation(self, read_data, read_var_list):!\label{env_get_observation}!   
        [...]

    # compute a reward signal
    def _get_reward(self):!\label{env_get_reward}!   
        [...]

    # close external resources used by environment's physics simulation engine
    def _close_external_resources(self):!\label{env_close}!
        [...]

    def step(self, action):  # overridden step function!\label{env_step}!
        # step through environment smoothly and linearly from old to new control action 
        [...]

    [...]  # private helper functions!\label{env_helpers}!
\end{minted}
\end{minipage}
\end{center}
\caption{The code excerpt of environment API for the synthetic jet flow control case.}
\label{list:2}
\end{listing}
\begin{listing}[!ht]
\begin{center}
\begin{minipage}{0.5\textwidth} 
\begin{minted}{json}
{
    "environment": {
        "name": "example_1"
    },
    "physics_simulation_engine": {
        "solvers": ["fluid-openfoam"],
        "reset_script": "reset.sh",
        "run_script": "run.sh"
    },
    "controller": {
        "read_from": {},
        "write_to": {
            "jet1": "Velocity",
            "jet2": "Velocity"
        }
    }
}
\end{minted}
\end{minipage}
\end{center}
\caption{Adapter configuration file, \mintinline{bash}{gymprecice-config.json}, for the synthetic jet flow control case.}
\label{list:3}
\end{listing}

Listing~\ref{list:3} shows the configuration file of the test case, \mintinline{bash}{gymprecice-config.json}. After converting the \mintinline{bash}{json} file to a Python dictionary, the resulting data is passed to the adapter as \mintinline{bash}{options} argument. This argument is used to configure the three main components of the control loop, which include the environment, controller, and physics simulation engine. The environment name is used as a label added to the output file names. The physics simulation engine for this test case has a single fluid-solver directory named \mintinline{bash}{fluid-openfoam}, which contains the OpenFOAM simulation case along with \mintinline{bash}{reset.sh} and \mintinline{bash}{run.sh} \mintinline{bash}{bash} scripts. The controller writes a surface field called \mintinline{bash}{Velocity} to \mintinline{bash}{jet1} and \mintinline{bash}{jet2} boundary patches (actuation interfaces) of the fluid-solver, without directly reading any data from the physics simulation engine.

\begin{figure}[t]
    \centering
    \includegraphics[scale=1.0]{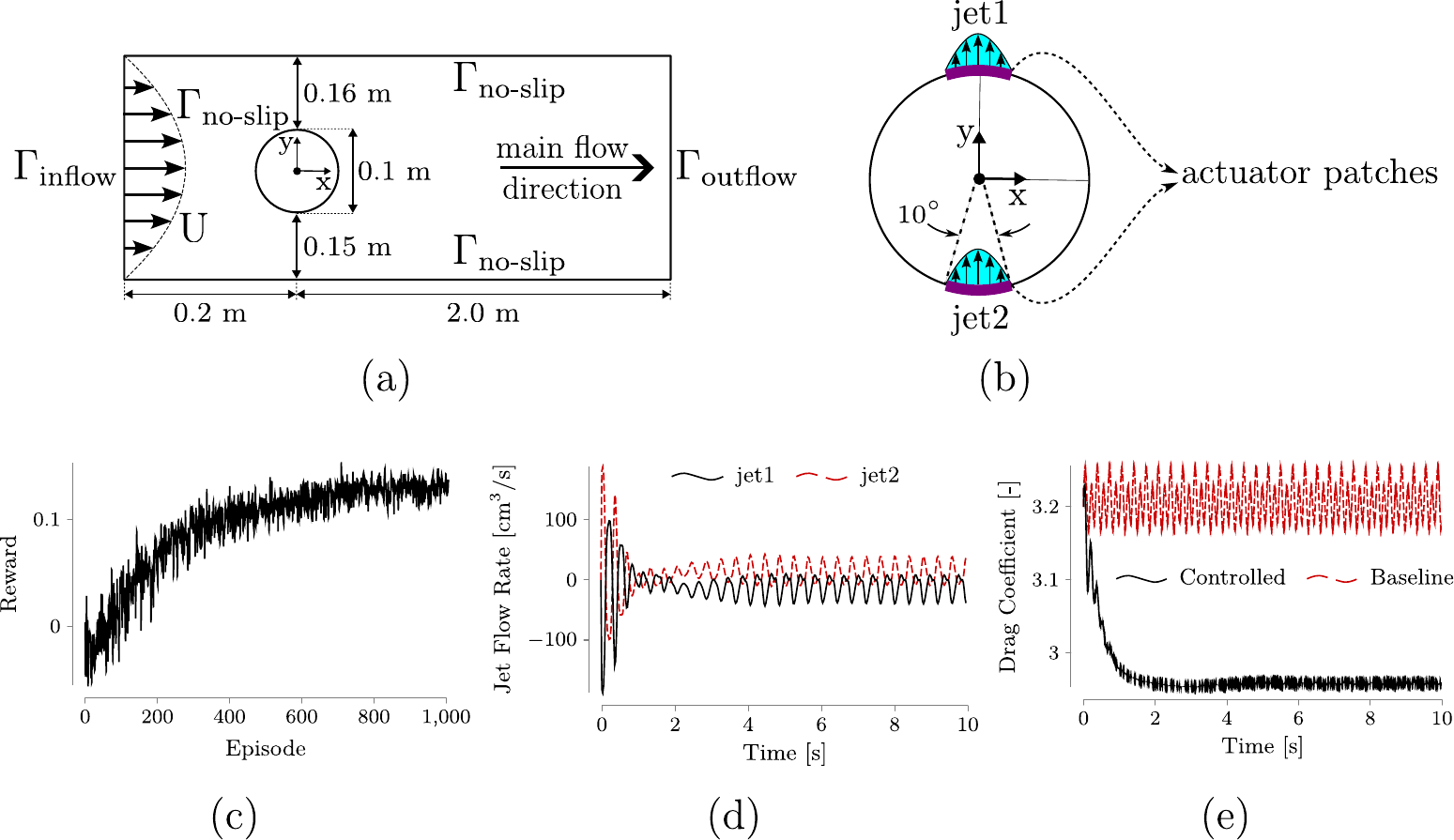}
    \caption{Synthetic jet flow control using a PPO agent. Schematics of (a) geometric description of the domain and boundary conditions used for simulating the flow over an immersed cylinder in a two-dimensional channel flow, and (b) synthetic jet actuator on the poles of the cylinder. Evolution profiles of (c) reward signal during training process of the PPO agent for 1000 episodes, (d) flow rate of the actuator controlled by a trained PPO agent, and (e) drag coefficient for controlled and baseline cases.}
    \label{fig:4}
\end{figure}

Figure~\ref{fig:4}a shows a schematic of the flow domain, with which the DRL agent interacts.  For the inlet, we prescribe a parabolic inflow profile with a maximum velocity of U$_{max}=1.5$~m/s. We set the fluid density to $\rho=1$~kg/m$^3$, and the kinematic viscosity to $\nu=0.001$~m$^2$/s. Figure~\ref{fig:4}b shows a schematic of the jets (actuation interfaces) with angular width of $10^\circ$ mounted symmetrically on the cylinder poles, injecting fluid normal to the cylinder surface. The jets are controlled by the DRL algorithm through a zero-net flow rate limited to a maximum value of $250.0$~cm$^3$/s. Figure~\ref{fig:4}c depicts the profile of the reward value throughout the multi-environment training process of the DRL controller for 1000 episodes. During the training, the PPO algorithm simultaneously interacts with 24 parallel environments, each of which a two-second fluid flow simulation (episode) and collects pressure data measured by 151 probes located within the flow field. The effectiveness of the trained DRL controller is tested for a control simulation lasting 10 seconds. As shown in Figure~\ref{fig:4}d, the temporal evolution of the control action becomes periodically stable. Figure~\ref{fig:4}e shows that the trained DRL controller successfully reduces drag coefficient, a dimensionless quantity used for measuring the drag force, by about $8\%$ compared to the baseline case with no jet flow control.

\subsubsection{Rotating cylinder flow control}
\begin{figure}[t]
    \centering
    \includegraphics[scale=1.0]{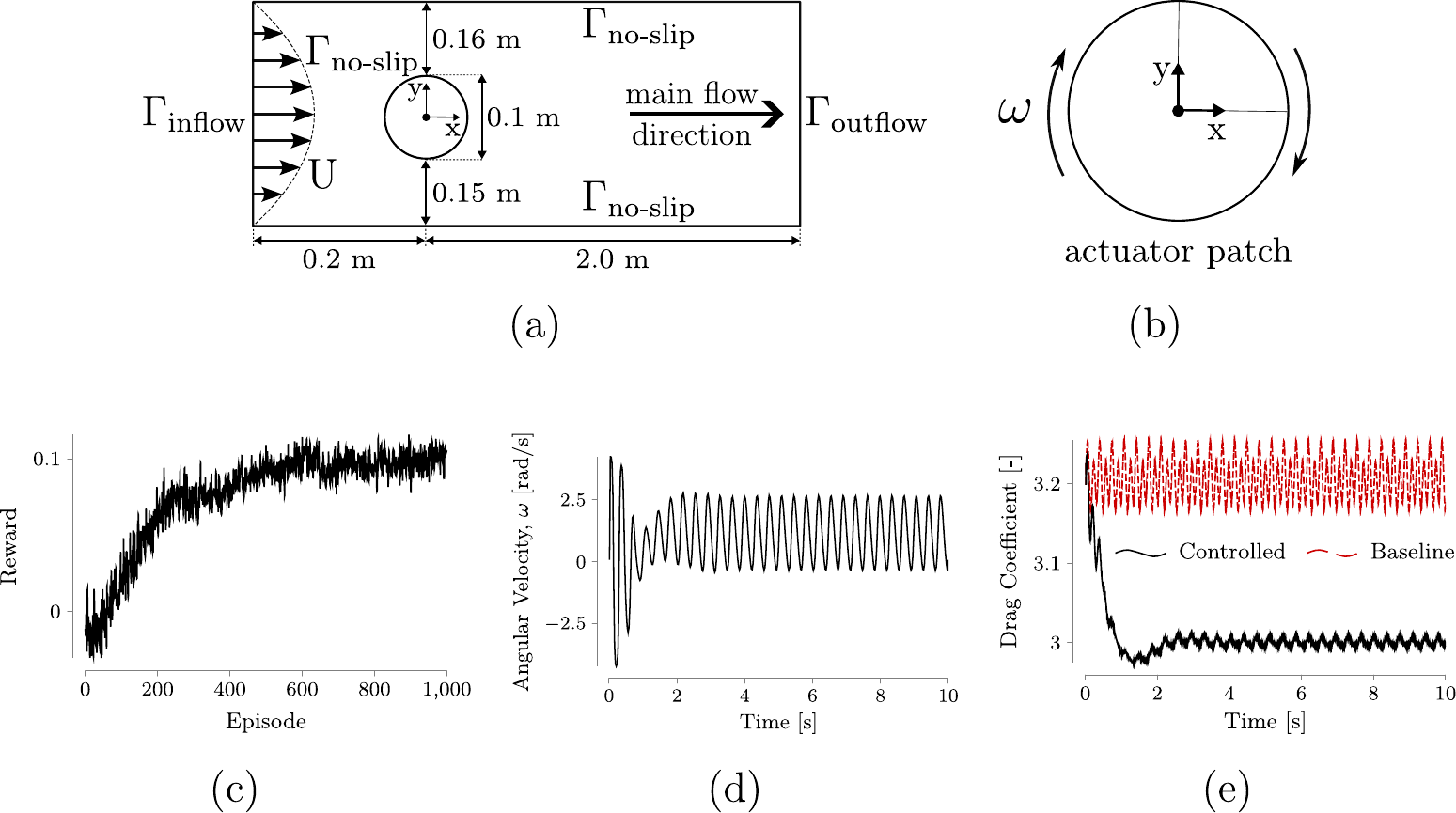}
    \caption{Rotating cylinder flow control using a PPO agent. Schematics of (a) geometric description of the domain and boundary conditions used for simulating the flow over an immersed cylinder in a two-dimensional channel flow, and (b) the cylinder as a rotating actuator with an angular velocity of $\omega$~(rad/s). Evolution profiles of (c) reward signal during training process of the PPO agent for 1000 episodes, (d) angular velocity of the actuator controlled by a trained PPO agent, and (e) drag coefficient for controlled and baseline cases.}
    \label{fig:5}
\end{figure}

To showcase the flexibility of our framework, we consider a modified version of the previous example in which the synthetic jet actuator is substituted with a rotating cylinder actuator\footnote{\url{https://github.com/gymprecice/tutorials/tree/v0.1.0/closed_loop_AFC/rotating_cylinder}}\textsuperscript{,}\footnote{The concept of utilising a rotating cylinder as the actuator is adapted from \url{https://github.com/OFDataCommittee/drlfoam/tree/main/openfoam/test_cases/rotatingCylinder2D}}. Figure~\ref{fig:5}a shows a schematic of the flow domain. Figure~\ref{fig:5}b shows a schematic of the rotating cylinder controlled by the DRL algorithm through an angular velocity limited to a maximum value of $\omega = 5$~rad/s. While the environment is nearly identical to the previous test case, the only key variation lies in the \mintinline{bash}{_get_action} method, which prescribes the velocity boundary field on the surface of the control cylinder to match the surface speed of a rotating cylinder with an angular velocity determined by the DRL controller. Figure~\ref{fig:5}c depicts the learning performance of the PPO agent. Figure~\ref{fig:5}d, shows the angular velocity of the rotating cylinder prescribed by the trained PPO agent for a control simulation lasting 10 seconds. Figure~\ref{fig:5}e shows the corresponding drag coefficient profile that corresponds to the same control period.

\subsection{Open-loop active flow control}
Our design software offers non-intrusive communication by utilising the preCICE library to interface middle software layers such as Gym-preCICE adapter, OpenFOAM-preCICE adapter, etc. This communication approach provides a remarkable level of flexibility and extensibility to define RL environments for various AFC problems, with only a few lines of Python code needed to define a problem-specific environment. In the following example, we illustrate the capability of our software in controlling a relatively more complex multi-solver physics simulation engine, in which a fluid-structure interaction (FSI) case is simulated using an OpenFOAM fluid solver and a deal.II solid solver.

\subsubsection{Fluid–structure interaction control}
\begin{figure}[t]
    \centering
    \includegraphics[scale=1.0]{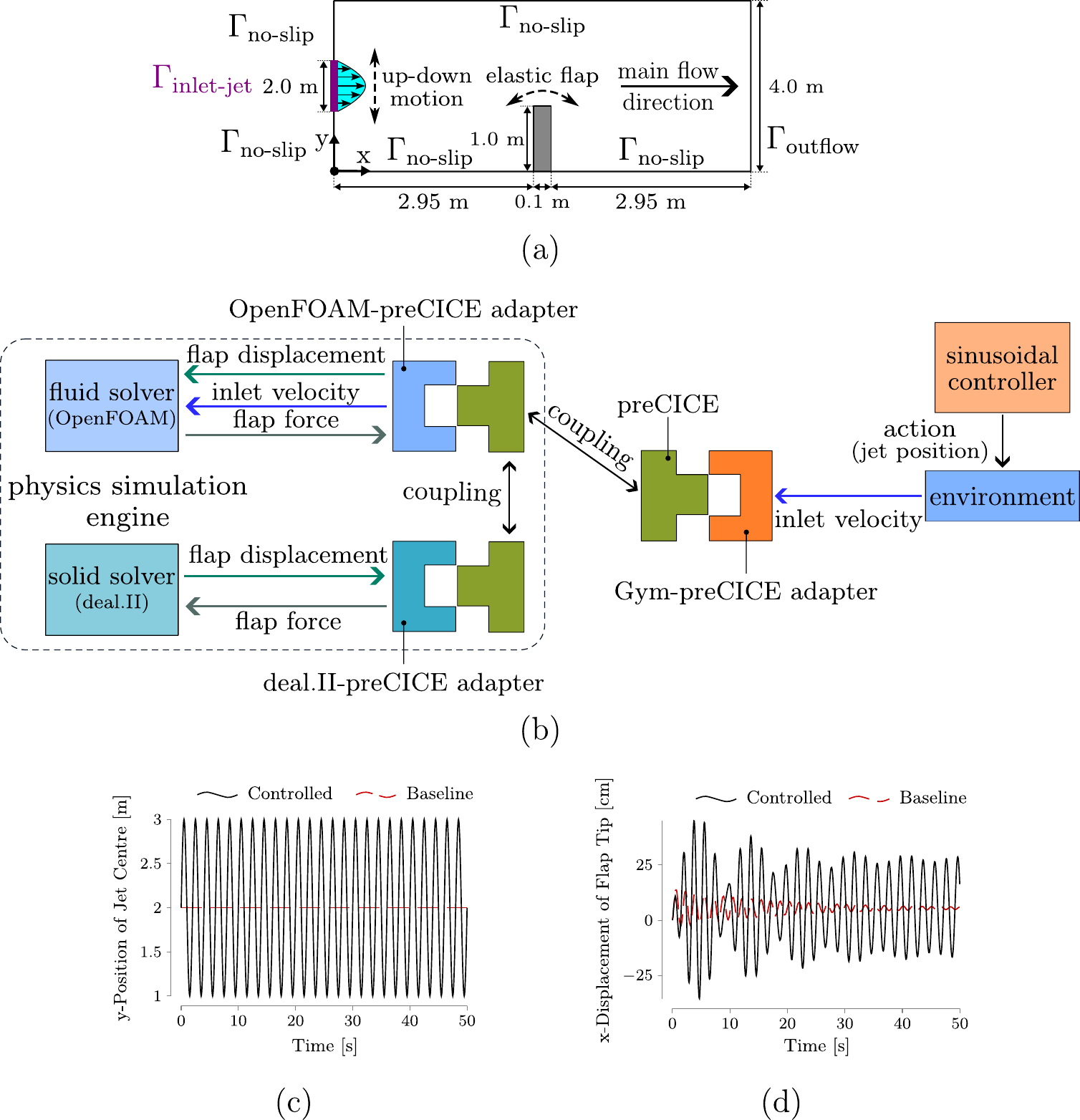}
    \caption{Fluid-structure interaction open-loop control using a predefined sinusoidal controller. Schematics of (a) geometric description of the domain and boundary conditions used for simulating the motion of a wall-mounted elastic flap in a two-dimensional channel flow, and (b) an open-loop control framework in which a predefined sinusoidal controller periodically changes the inflow profile of the channel by controlling the position of a jet at the inlet. Profiles of (c) the control action over time, and (d) the tip displacement of the elastic flap in x-direction over time, for controlled and baseline cases.}
    \label{fig:6}
\end{figure}

As a simple FSI control test case, we consider manipulating the motion of a wall-mounted elastic flap in a two-dimensional channel flow\footnote{\url{https://github.com/gymprecice/tutorials/tree/v0.1.0/open_loop_AFC/perpendicular_flap}}\textsuperscript{,}\footnote{The physics simulation engine for this test case is adapted from \url{https://github.com/precice/tutorials/tree/v202211.0/perpendicular-flap}}. We use a predefined sinusoidal controller to control the centre position of a jet along the channel inlet. Figure~\ref{fig:6}a shows a schematic of the FSI domain. For the jet, we prescribe a parabolic inflow profile with a maximum velocity of U$_{max}=15.0$~m/s. We set the fluid density to $\rho_{f}=1.0$~kg/m$^3$, the fluid kinematic viscosity to $\nu_{f}=1.0$~m$^2$/s, the solid density to $\rho_{s}=3.0 \times 10^{3}$~kg/m$^3$, the Young’s modulus to $E=4.0$ MPa, and the Poisson ratio to $\nu_{s}=0.3$. Figure~\ref{fig:6}b schematically depicts the open-loop control framework. The process of the control loop is as follows: (1) the FSI environment receives the jet centre y-position from the controller, (2) the environment, based on the jet location, prescribes a velocity boundary field for the inlet of the channel, and passes the boundary field to Gym-preCICE adapter, (3) the adapter writes the boundary field to preCICE, and (4) preCICE communicate the boundary values with the physics simulation engine of the FSI setup. Figure~\ref{fig:6}c and Figure~\ref{fig:6}d depict the profile of the control action and its impact on the tip displacement of the elastic flap in x-direction over time.

%% file: sections/impact.tex
The Gym–preCICE adapter couples reinforcement learning algorithms and mesh-based PDE solvers in a nearly plug-and-play fashion. This can dramatically simplify the usage of reinforcement learning in the field of AFC. Particularly, the adapter provides a flexible, modular, and easy-to-use framework to perform quick comparative assessment of different reinforcement learning algorithms for an AFC problem, and to devise highly adaptable, fast, and accurate intelligent control algorithms for fluid flow systems. By enabling reproducibility and lowering the barrier-to-entry, we believe, the Gym–preCICE will motivate interdisciplinary research at the interface of machine learning and computational mechanics to address a vast class of research questions in AFC. The Gym–preCICE adapter is distributed under the MIT license, which allows both academic and commercial application of the adapter while encouraging contributions from all researchers to its future versions. Given the popularity of the PDE solvers supported by preCICE, such as OpenFOAM, deal.II, FEniCS, etc., both in academia and industry, the rapid growth of preCICE users in a wide variety of application areas, and the recent surge in developing Gym-style RL libraries, we expect significant interest in the adapter in the coming years.

%% file: sections/conclusions.tex
In this paper, we presented the new software Gym–preCICE, a Gymnasium-style interface written in Python that allows the coupling of RL algorithms and numerical PDE solvers via preCICE. We demonstrated the use of our work for multi-environment training of a DRL agent to reduce drag in two-dimensional incompressible fluid flow over a cylinder simulated using OpenFOAM library. The agent was able to learn to attenuate the drag in two different AFC scenarios: (1) controlling the flow rate of a synthetic jet actuator installed on a stationary cylinder, and (2) controlling the angular velocity of a rotating cylinder. By employing the framework to exert control in a fluid-structure interaction setup, we showed that Gym-preCICE is generic enough to accommodate more than one PDE solver (from different simulation software packages) within a control loop.

The impact of the new software should be significant in bridging the gap between reinforcement learning and active flow control research. Gym-preCICE adapter allows users to couple PDE solvers to state-of-the-art DRL libraries by only changing a few lines of code in a problem-specific Python class and use it as the environment of a DRL controller. This should accelerate the scientific innovation in active flow control research by lowering software development and coupling complexity that has limited the wide adoption of DRL in active flow control. In future, we aim to extend the applicability of the Gym-preCICE adapter to a broader set of research challenges that involve fluid-fluid interaction, fluid-structure interaction, and conjugate heat transfer. We will put the necessary efforts into documentation, tutorials, packaging, testing, and integration with other simulation software packages supported by preCICE. This will help grow the user-base and form an interdisciplinary knowledge-sharing and collaboration network around Gym–preCICE.